\pdfoutput=1

\documentclass[11pt]{article}

\usepackage{EMNLP2023}

\usepackage{times}
\usepackage{latexsym}

\usepackage[T1]{fontenc}
\usepackage{graphicx}
\usepackage{float}
\usepackage{stfloats}
\usepackage{color,soul}
\usepackage[utf8]{inputenc}
\usepackage{pdflscape}
\usepackage{lscape}
\usepackage{longtable}
\usepackage[toc,page]{appendix}
\usepackage[parfill]{parskip}
\usepackage{mathtools}
\usepackage{amsmath}
\usepackage{multirow}
\usepackage{tabularx}
\usepackage{colortbl}

\usepackage{microtype}

\usepackage{inconsolata}
\usepackage{fancyhdr}
\newcommand{\mypm}{\mathbin{\smash{%
\raisebox{0.35ex}{%
            $\underset{\raisebox{0.5ex}{$\smash -$}}{\smash+}$%
            }%
        }%
    }%
}

\title{How well ChatGPT understand Malaysian English? An Evaluation on Named Entity Recognition and Relation Extraction}

\author{Mohan Raj Chanthran$^1$, Lay-Ki Soon$^{1*}$, Huey Fang Ong$^1$, and Bhawani Selvaretnam$^2$
       \\
       $^1$School of Information Technology, Monash University Malaysia\\
       \{mohan.chanthran, soon.layki, ong.hueyfang\}@monash.edu\\ 
       $^2$Valiantlytix\\
       {bhawani@valiantlytix.com} \\
       }

\graphicspath{ {./Figure/} }

\begin{document}
\maketitle
\begin{abstract}
Recently, ChatGPT has attracted a lot of interest from both researchers and the general public. While the performance of ChatGPT in named entity recognition and relation extraction from Standard English texts is satisfactory, it remains to be seen if it can perform similarly for Malaysian English. Malaysian English is unique as it exhibits morphosyntactic and semantical adaptation from local contexts. In this study, we assess ChatGPT's capability in extracting entities and relations from the Malaysian English News (MEN) Dataset. We propose a three-step methodology referred to as \textbf{\textit{educate-predict-evaluate}}. The performance of ChatGPT is assessed using F1-Score across 18 unique prompt settings, which were carefully engineered for a comprehensive review. From our evaluation, we found that ChatGPT does not perform well in extracting entities from Malaysian English news articles, with the highest F1-Score of 0.497. Further analysis shows that the morphosyntactic adaptation in Malaysian English caused the limitation. However, interestingly, this morphosyntactic adaptation does not impact the performance of ChatGPT for relation extraction. 

\def\thefootnote{*}\footnotetext{Corresponding Author.}\def\thefootnote{\arabic{footnote}}
\end{abstract}

\section{Introduction}
\label{sec:introduction}

With the recent emergence of Large Language Models (LLM), we have observed a paradigm shift in natural language processing (NLP). These LLM include PaLM \citep{chowdhery2022palm}, ChatGPT \citep{openai_gpt3.5}, GPT-4 \citep{openai_gpt4}, and Llama 2 \citep{touvron2023llama}. ChatGPT, in-arguably the most popular LLM currently, is developed by OpenAI and has demonstrated remarkable ability in language understanding and generating coherent responses. The use of ChatGPT has been observed in various NLP tasks, including Sentiment Analysis \citep{wang2023chatgpt, belal2023leveraging}, Topic Classification \citep{reiss2023testing,gilardi2023chatgpt}, and Information Extraction \citep{wei2023zeroshot,li2023evaluating,hu2023zeroshot}. There have been several research works conducted to evaluate the capabilities of ChatGPT for NER and RE \citep{li2023evaluating, han2023information, chan2023chatgpt}. While most of the evaluation outcomes focused on Standard English, it raises a question: \textit{Is ChatGPT capable of extracting entities and relations from Malaysian English News}? 

Originating from Standard English, Malaysian English (ME) has evolved into a unique form of English incorporating local words from languages like Bahasa Malaysia, Chinese and Tamil \citep{malaysian-english-versus-standard-english}. Malaysian English exhibits usage of Loan Words, Compound Blends and Derived Words \citep{exploring-malaysian-english-newspaper-corpus}. Some example sentences with the usage of Loan Words, Compound Blends and Derived Words are provided, such as:
\begin{enumerate}
    \item "... billion of jobs in the next five to seven years, as well as Bukit Bintang City Centre with RM600 million jobs awarded so far". From this sentence, Bukit Bintang City Centre is a compound blend where "Bukit Bintang" refers to the name of LOCATION in Bahasa Malaysia, and this entity refers to a shopping mall (LOCATION).
    \item "... economy to provide higher-paying jobs in cutting-edge technology for Selangorians, he said". From this sentence, "Selangorians" is a derived word that indicates the people from the state of Selangor.    
    \item "KUALA LUMPUR: Prime Minister Datuk Seri Anwar Ibrahim today urged ... business tycoon Tan Sri Syed Mokhtar Albukhary ...". From this sentence, "Datuk Seri" and "Tan Sri" is a loanword, it is a common honorific title given for PERSON.
\end{enumerate}
The existence of loan words, compound blends, and derived words in the usage of entity mentions has motivated us to assess the performance of ChatGPT in Malaysian English, specifically for Named Entity Recognition (NER) and Relation Extraction (RE).


Prompting techniques like Zero Shot, Few Shot, and Chain of Thought (CoT) have been proven to improve the performance of ChatGPT in various NLP tasks \citep{brown2020language, han2023information, chan2023chatgpt, wei2023chainofthought}. In-context learning helps ChatGPT to understand more about the task in hand and define the scope on the task to be completed. It has been proven effective for domain-specific tasks, such as legal reasoning \citep{kang2023can}.  Keeping these in mind, we propose a novel three-step method to extract entities and relations from Malaysian English news articles, called "\textit{educate-predict-validate}". Section \ref{sec:educate-predict-evaluate} discusses these three steps in detail.

ChatGPT's ability to extract entities and relations is measured based on its agreement with human-annotated labels using the F1-Score. Our evaluation aims to establish a benchmark for ChatGPT's performance in Malaysian English texts. The code for this experiment is available at Github\footnote{\url{https://github.com/mohanraj-nlp/ChatGPT-Malaysian-English}} for reproducibility. The contributions of this research can be summarised as follows:
\begin{enumerate}
    \item \textit{In-context learning for better ChatGPT performance}.  A novel approach to identify and extract entities and relations from any document or text by providing sufficient contexts to ChatGPT. 
    \item \textit{Comprehensive assessment of ChatGPT performance on Malaysian English News Articles}. A total of 18 different prompt settings have been carefully engineered to evaluate ChatGPT's capability in NER and RE. 
    The output produced by ChatGPT is compared against human-annotations. 
\end{enumerate}

In short, the analyses reported in this paper answer these questions: a) \textit{How well does ChatGPT perform in extracting entities from Malaysian English?}; b) \textit{Are there specific types of entity labels that ChatGPT consistently struggle to extract or misidentified?}; c) \textit{How accurate is ChatGPT in extracting relations between entities?}; d) \textit{How good is ChatGPT in predicting entities and relation from Standard English?}.

Section \ref{sec:related-work} presents the evaluation done on ChatGPT for Standard English.   Section \ref{sec:educate-predict-evaluate} discusses our proposed "\textit{educate-predict-validate}" methodology.  Section \ref{sec:experiment} describes our experimental setup. Section \ref{sec:result-analysis} presents our experiment results and findings, including an analysis of the challenges and limitations encountered by ChatGPT when handling Malaysian English news articles. Finally in Section \ref{sec:conclusion} we have concluded our work and our future work.

\section{Related Work}
\label{sec:related-work}

\subsection{LLM for Information Extraction}
\label{ssec:chatgpt-ie}
To understand the capabilities of LLM on entity and relation extraction, we have gone through some recent research on LLM for Information Extraction (IE). \citep{wei2023zeroshot} has proposed ChatIE, a zero-shot information extraction framework using ChatGPT. The information extraction task will be conducted into two stages and it will be based on question-answering approach. In the first stage, a sentence will be passed to ChatGPT followed by a question asking whether the sentence contains any entities, relations, or event types from a predefined list. The question prompt will include the list of entity, relation, or event types. In the second stage, the prompt will be modified depending on the specific task. For NER, the entity type extracted from first stage will be given to ChatGPT to extract all entity mentions. Meanwhile, for RE, both entity type and relation type will be given to ChatGPT to identify entity mentions that match with the entity type and relation. ChatIE improves performance by an average of 18.98\% compared to ChatGPT without ChatIE. However it is noticeable that the F1-score varies depend on the dataset that has been tested upon. 

\citep{li2023evaluating} assesses the ability of ChatGPT in 7 Fine Grained IE tasks like Entity Typing, NER, Relation Classification, and RE. The prompt is formulated by considering two distinct configurations: Standard-IE settings and OpenIE settings. Compared to the baseline and SOTA models, ChatGPT's performance is less competent. For NER tasks, ChatGPT performance is lower in OntoNotes (with 18 Labels) compared to ConLL (4 Labels). For relation classification and RE, ChatGPT performance is lower in TACRED (42 Labels) compared to SemEval2010 (10 Labels).

\citep{han2023information} conducted an extensive evaluation to examine the performance of ChatGPT in IE. A total of 14 subtasks related to IE were tested using 17 distinct datasets. The experimental conditions employed in this study encompass three prompt settings: zero shot prompt, few shot prompt, and few shot with CoT prompts. The experiments conducted evaluated several subtasks that are relevant to our research, including NER-Flat, NER-Nested, Relation Triplet (RE-Triplet), and Relation Classification (RE-RC). The experimental results showed that ChatGPT exhibited superior performance in the NER-Flat task as compared to the NER-Nested task. The F1-Score for RC-RE reached its lowest value at 19.47 when evaluated on the TACRED dataset under zero shot conditions. In the case of RE-Triplet, the dataset NYT-multi exhibited the lowest F1-Score, which amounted to 3.45. The experimental results also indicated that ChatGPT did poorly in relation classification for entities, with its lowest performance observed in triplet extraction. 

\begin{figure*}[!ht]
  \includegraphics[height=\textheight,width=\textwidth,keepaspectratio]{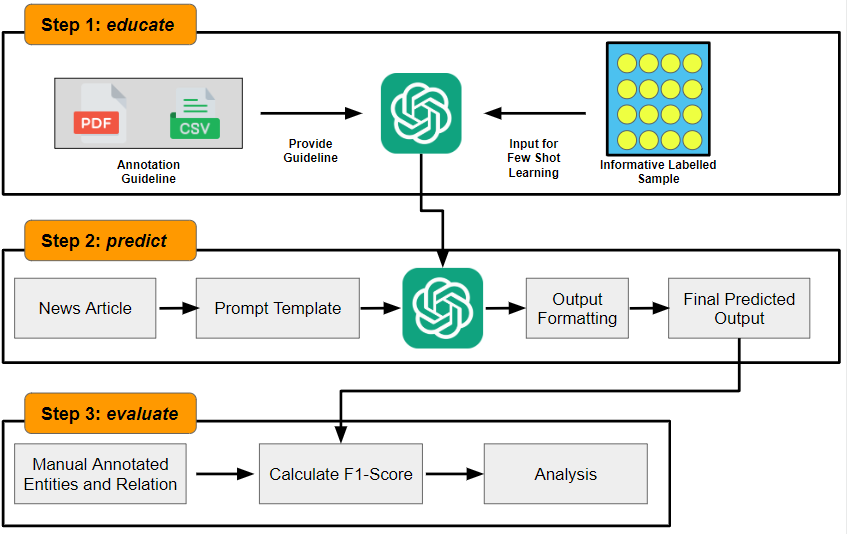}
  \caption{Detailed steps in the proposed \textit{educate-predict-evaluate} methodology}
  \label{fig:educate-predict-evaluate}
\end{figure*}

\section{\textit{educate-predict-evaluate}}
\label{sec:educate-predict-evaluate}

ChatGPT is one of the widely used Large Language Models. It can be easily interacted through the provided Web interface, by asking questions and make conversation with the model. Providing additional context helps ChatGPT to learn and better understand the tasks in hand. In this paper, we propose a systematic methodology called \textit{educate-predict-evaluate}, which aims to carry out a comprehensive evaluation on ChatGPT capability in NER and RE within Malaysian English context. Figure \ref{fig:educate-predict-evaluate} shows detailed view of proposed approach. 
\begin{enumerate}
    \item \textit{educate}: The idea behind this is to teach ChatGPT how to extract entity and relation from Malaysian English texts. To accomplish this, we provided ChatGPT with the annotation guideline prepared while developing MEN-Dataset \citep{chanthran2024malaysian}. This approach is also called as In-Context Learning (ICL). Appendix \ref{appendix:icl-annotation} shows a sample of prompt generated with the annotation guideline for extracting entities. Apart from guideline, we also applied Few Shot Learning approach. In Few Shot Learning, we provided a few news articles with annotated entities and relations. In addition, we also provided some explanations that include the context, or justifications on why entities and relations are extracted from news article. These explanations were provided by the human annotators who contributed to developing and annotating MEN-Dataset. Appendix \ref{appendix:entities-explainations} presents some samples of explanations given for entity extraction. 
    \item \textit{predict}: We propose a Self-Consistent Few Shot Prompting Technique, together with the explanation on why each entity has been annotated by the human annotator. The explanation acts as additional context for ChatGPT to identify the entities and relations. \citep{wang2022selfconsistency} proposed the Self-Consistent prompting techniques, where the idea behind is to choose the most consistent answer as the final answer of ChatGPT. For instance, a prompt for a chosen news article will be provided to ChatGPT three times, and the entities that have been extracted more than twice will be considered as final output for the particular news article. In Table \ref{tab:prompt-ner-re}, we have listed all 18 different prompt settings used in this experiment. Appendix \ref{appendix:ner-task} presents the prompt used to extract entities while Appendix \ref{appendix:re-task} presents the prompt used to identify relations from news articles. 
    \item \textit{validate}: We have assessed the performance of ChatGPT on NER and RE by calculating the F1-Score with human annotation provided by the dataset.
\end{enumerate}

\section{Experiment}
\label{sec:experiment}

\subsection{Dataset}
\label{ssec:dataset}
\begin{table}[]
\centering
\begin{tabular}{|l|l|}
\hline
Statistics                                                                       & Frequency \\ \hline
Total Entities                                                                   & 6,061      \\ \hline
Total Unique Entities                                                            & 2,874      \\ \hline
Total Relations                                                                  & 3,268      \\ \hline
\begin{tabular}[c]{@{}l@{}}Total Relation based \\ on DocRED Labels\end{tabular} & 2,237      \\ \hline
\begin{tabular}[c]{@{}l@{}}Total Relation based \\ on ACE-2005 Labels\end{tabular} & 1,031      \\ \hline
\end{tabular}
\caption{The statistics of total Entities and Relation annotated in MEN}
\label{tab:entity-relation-statistics}
\end{table}
We used two datasets to evaluate the performance of ChatGPT for NER and RE, which include:
\begin{enumerate}
    \item MEN-Dataset is a Malaysian English news article dataset with annotated entities and relations \citep{chanthran2024malaysian}. We have built the dataset with 200 news articles extracted from prominent Malaysian English news article portals such as New Straits Times (NST)\footnote{\url{https://www.nst.com.my/}},  Malay Mail (MM)\footnote{\url{https://www.malaymail.com/}} and Bernama English\footnote{\url{https://www.bernama.com/en/}}. The dataset consists of 12 entity labels, and 101 relation labels. Appendix \ref{appendix:men-dataset-entity-labels} and Appendix \ref{appendix:men-dataset-relation-labels} contain the complete lists of entity and relation labels, respectively. For entities, we have adapted the labels from the OntoNotes 5.0 dataset \citep{ontonotes-new}. The relation labels are adapted from ACE05 \citep{Walker2005-ym} and DocRED \citep{yao-etal-2019-docred}. Table \ref{tab:entity-relation-statistics} presents the statistics of the entities and relations annotated in the dataset.  
    \item DocRED: DocRED \citep{yao-etal-2019-docred} is a prominent dataset designed specifically for inter-sentential relation extraction models. The dataset includes annotated entities and relations. The dataset has been chosen to facilitate a comparative analysis of ChatGPT's performance in both Malaysian English and Standard English. 
\end{enumerate}

While we have adapted the entity labels from OntoNotes 5.0 and the relation labels from ACE-2005, we did not use these datasets for this evaluation. The OntoNotes 5.0 dataset is structured at the sentence level, with entity annotations specific to each individual sentence. An earlier effort showed that ChatGPT does not perform well on longer text \citep{han2023information}. To mitigate the impact of input length on ChatGPT's performance, we have opted to utilize a dataset containing longer context sequences. This decision led us to select DocRED for evaluation. It is also important to note that the MEN dataset encompasses both inter and intra-sentential relations.

\subsection{Experimental Setup}
\label{ssec:experimental-setup}
The experiment was conducted in between April 2023 and August 2023. Notably, the outcome of ChatGPT exhibited variability over time \citep{chen2023chatgpts}.  While OpenAI API is available, we decided to use ChatGPT\footnote{\url{https://chat.openai.com/}} official website. There were several reasons for our decision, and these have been discussed in Section \ref{sec:limitation}.
To ensure a fair comparison, we used 195 articles for experiment. Another five articles were used for Few-Shot learning context. The In-Context Learning technique involves the integration of annotation guidelines and/or a limited set of few-shot samples as input of ChatGPT. During the process of picking few-shot samples, we implemented a filtering mechanism to identify and prioritize samples that possess the highest quantity of annotated entities or relation labels. For NER, we provided articles as input; meanwhile, for RE, we provided articles and entity pairs. For the evaluation metrics, we utilized F1-Score, and Human Validation, as mentioned in Section \ref{sec:result-analysis}. The F1-Scores were calculated by comparing ChatGPT's predictions with human annotations in the dataset.

\begin{figure*}[!ht]
  \centering
  \includegraphics[height=\textheight,width=\textwidth,keepaspectratio]{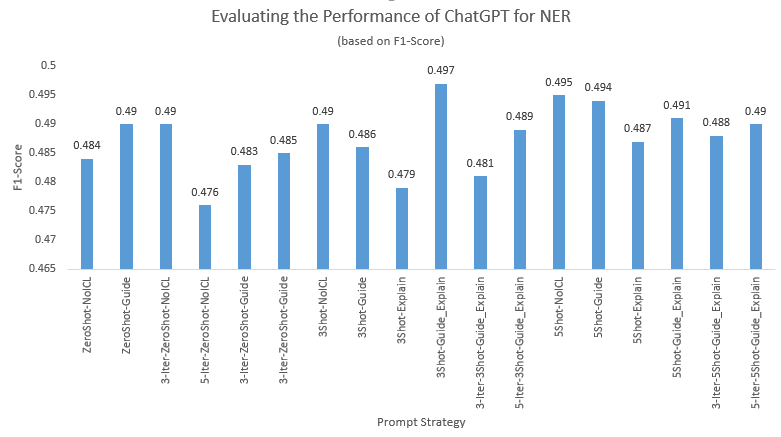}
  \caption{F1-Scores based on entities extracted by ChatGPT for Malaysian English news articles.}
  \label{fig:men-dataset-f1-chatgpt}
\end{figure*}
\section{Result and Analysis} 
\label{sec:result-analysis}

In this section, we present the outcome of the experiment that we conducted. In Section \ref{ssec:eval-chatgpt-on-men-dataset}, we discuss how ChatGPT performs NER and RE on MEN-Dataset, together with the observed limitations. 

\subsection{\textit{How well did ChatGPT perform in extracting entities from Malaysian English? Does it perform better?}}
\label{ssec:eval-chatgpt-on-men-dataset}
Figure \ref{fig:men-dataset-f1-chatgpt} shows the experiment results using different prompt settings. Some observation made from Figure \ref{fig:men-dataset-f1-chatgpt} are:
\begin{enumerate}
    \item ChatGPT achieved highest F1-Score with prompt 3 Shot+Guideline+Explanation. From the overall experiment, the average F1-Score recorded was 0.488, and the highest F1-Score was 0.497. The result shows that providing a few shot samples with explanation and annotation guidelines enabled ChatGPT to do NER by complying with the instructions. Providing three-shot samples with annotation guidelines was sufficient for ChatGPT to understand the task and annotate. 
    \item The impact of the guidelines is significant in improving the performance of ChatGPT. Each non-consistent prompt technique with guidelines improved the performance of ChatGPT in comparison to outcome without guidelines.  
    \item Self-consistent technique is not effective in ensuring quality output by ChatGPT. If we compare the experiment results with and without self-consistent approach for zero-shot, the F1-Score with the self-consistent approach is lower. This shows that integrating the Self-Consistent technique with few shot learning approaches did not yield substantial improvements in all cases. However, this technique helps to ensure the consistency of the outcome. 
    \item Although we made multiple prompting strategies, the overall F1-score did not improve significantly. The overall difference of F1-Score recorded is 0.488 $\mypm$ 0.01.
\end{enumerate}
During the annotation of the MEN-Dataset, we calculated the Inter-Annotator Agreement (IAA) using the F1-Score and achieved a score of 0.81. Meanwhile, the highest F1-Score achieved by ChatGPT from this experiment was 0.497. This shows that there are still some limitations that can be observed from ChatGPT.

\subsection{\textit{What are the limitations of ChatGPT in extracting entities? Were there specific types of entity labels that ChatGPT consistently struggled to extract or misidentify?}}
\label{ssec:eval-chatgpt-on-men-dataset-entity-label} 
In Table \ref{tab:f1-score-entity-label}, we can see the F1-Score from the perspective of entity label level. This helps us to understand more about how ChatGPT extracts the entities. We manually checked the outcome from ChatGPT to understand its limitation in extracting entities. The following findings were observed from the outcomes generated by self-consistent prompting:
\begin{enumerate}
    \item Entity labels like PERSON, LOCATION, and ORGANIZATION have more than 1000 entity mentions annotated in MEN-Dataset. While the remaining entity labels have a total entity mention of less than 300. 
    \item The entity label PERSON has an average F1-Score of 0.507. Our analysis noticed that most errors happened due to Loan Words and Compound Blend found in Malaysian English news articles. Here are some examples:
    \begin{enumerate}
        \item Tan Sri Dr Noor Hisham Abdullah. "Tan Sri" is a loanword, a common honorific title for PERSON. It is often used to mention important personals. It is often used together with the name of PERSON.  
        \item Datuk Seri Haji Amirudin bin Shari. "Datuk Seri" is a loanword, a common honorific title for PERSON.  
    \end{enumerate}
    Apart from the errors due to Loan Words and Compound Blend, ChatGPT did not extract any co-referring entities. For example, \textit{Tan Sri Dr Noor Hisham Abdullah} is also used as \textit{Noor Hisham Abdullah} in a similar article, but ChatGPT did not extract it. 
    \item For ORGANIZATION, we noticed the importance of providing annotation guidelines. Several entity mentions from ORGANIZATION were not extracted before including the guideline in the prompts. Examples of entity mention are: \textit{Session Court, Public Mutual Funds, Parliment}. Furthermore, ChatGPT did not extract any abbreviations of entity mentions from entity label ORGANIZATION. Some examples: 
    \begin{enumerate}
        \item \textit{ATM}: The full form of ATM is "Angkatan Tentera Malaysia".
        \item \textit{Armada}: The full form of Armada is "Angkatan Bersatu Anak Muda".
        \item \textit{PN}: The full form of PN is "Perikatan Nasional".
    \end{enumerate}
    Similar issues observed for PERSON, where the co-reference of entity mentions was not extracted. 
    \item For NORP, we noticed most of the errors were due to \textit{Derived Words}. For instance, \textit{Sarawakians}, and \textit{Indonesian}.  The guideline included some examples for NORP, covering some frequently mentioned NORP, such as \textit{Bumiputera, Non-Bumiputera} and \textit{Malaysians}. The given examples were extracted correctly by ChatGPT. Apart from that, entity mentions with Loan Words like \textit{1998 Reformasi movement} were not identified by ChatGPT correctly. 
    \item Most of the entities mentioned from FACILITY that were not extracted by ChatGPT are with characteristics Compound Blend. The entities mentions from FACILITY have both English and Bahasa Malaysia, such as \textit{CIMB Bank Jalan Sagunting, Dataran Rakyat} and \textit{Aulong Sports Arena}. In addition, ChatGPT misidentified some entity labels. For instance, the entity mentioned that was supposed to be predicted as FACILITY was mistaken as LOCATION, and vice versa. Some other examples:
    \begin{enumerate}
        \item \textit{Kuala Lumpur International Airport} should be labeled as FACILITY instead of LOCATION.
        \item \textit{Jalan Langgak Golf} should be labeled as LOCATION instead of FACILITY.
        \item \textit{Sibujaya public library} should be labeled as FACILITY instead of LOCATION.
    \end{enumerate}  
    \item Most of the entity mentions in WORK\_OF\_ART are based on local creative works, consisting of Compound Blend. Some examples are \textit{Aku Mau Skola} and \textit{Puteri Gunung Ledang}.
    \item TITLE always appears together with the name of PERSON. In MEN-Dataset, the TITLE is annotated separately. The TITLE can be honorific or academic title. The honorific title consists of Loan Words like \textit{Datuk, Datuk Seri, Datin, Tan Sri} and more. 
\end{enumerate}

In conclusion, ChatGPT did not work well in extracting entity mentions with Loan Words, Compound Blend, and Derived Words. Apart from that, ChatGPT did not extract any co-reference entity mentions. Furthermore, any abbreviations of entity mentions were also not extracted by ChatGPT. 

\subsection{\textit{How accurate was ChatGPT in extracting relations between entities, and were there any notable errors or challenges?}}
\label{ssec:eval-chatgpt-on-men-dataset-relation}
The MEN-Dataset was annotated based on the relation labels adapted from DocRED and ACE05. There is also a special relation label named NO\_RELATION, which is annotated when no suitable relation labels exist for a particular entity pair. Due to the different characteristics of relation labels, we experimented with relation labels adapted from DocRED and ACE05 separately. We used prompt settings similar to the previous experiment.

\begin{figure*}[!ht]
  \centering
  \includegraphics[height=\textheight,width=\textwidth,keepaspectratio]{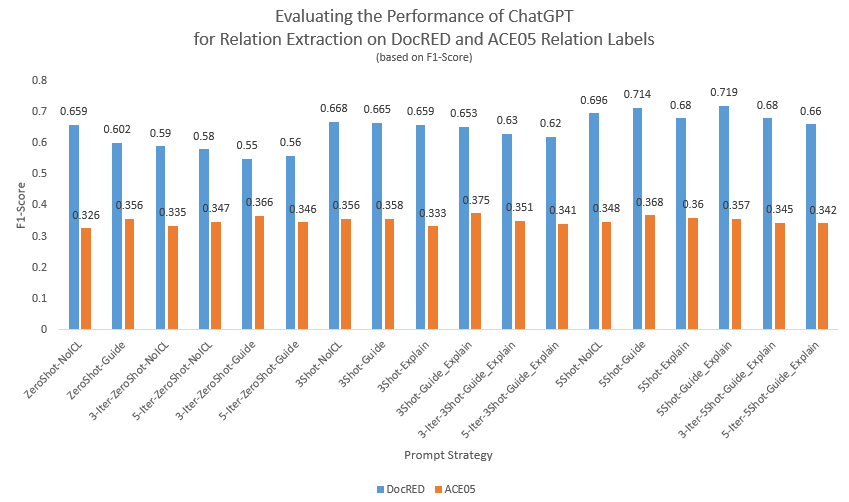}
  \caption{Performance of ChatGPT in classifying relations based on relation labels adapted from DocRED and ACE05}
  \label{fig:men-dataset-f1-chatgpt-docred-and-ace05}
\end{figure*}


Figure \ref{fig:men-dataset-f1-chatgpt-docred-and-ace05} shows the F1-Scores calculated based on the relations classified by ChatGPT for every entity pair. The average F1-Score for relation adapted from DocRED and ACE05 are 0.64 and 0.35 respectively. Some findings based on the results presented in Figure \ref{fig:men-dataset-f1-chatgpt-docred-and-ace05} are:
\begin{enumerate}
    \item \textbf{In-Context Learning improved the performance of ChatGPT in identifying the relations}. In both zero-shot and few-shot scenarios, the performance of ChatGPT has improved when providing both guidelines and explanations. 
    \item \textbf{Explanations made limited impact.} Including explanations and a few shot samples does not improve this task's performance. This approach has somehow improved the performance of ChatGPT in extracting entities. 
    \item \textbf{5 Shot Learning slightly improved the performance of ChatGPT, compared to 3 Shot Learning of various prompting techniques.} 
    \item \textbf{Complexity of relation labels.} When comparing the performance of ChatGPT across the two datasets, it is evident that the DocRED dataset produces a higher F1-Score than the ACE-2005 dataset. This can be seen across all evaluated prompting techniques. 
\end{enumerate}
One interesting observation is that in MEN-Dataset, 20\% of the relation triplets were labeled with NO\_RELATION. However, ChatGPT labeled as high as 80\% of the relation triplets as NO\_RELATION. While no morphosyntactical adaptation is involved when predicting the relation, understanding the context of the news article will impact the performance of ChatGPT in predicting the relations. 
In conclusion, we have seen the gap of ChatGPT on RE task for Malaysian English news article. To better understand the gap between Malaysian English and the Standard English, another question that may arise is \textit{How good is ChatGPT in NER and RE on Standard English?}

\subsection{\textit{How good is ChatGPT in predicting entities and relations from Standard English articles?}}
\label{ssec:predicting-entities-relation-standard-english}
In this experiment, we chose 195 articles with annotated entities and relations from DocRED. To ensure a valid comparison, we highlight some differences between MEN-Dataset and DocRED as follows: 
\begin{enumerate}
    \item In MEN-Dataset, we have 11 entity labels, while in the DocRED dataset, there are six entity labels. The overlapping entity labels are PERSON, ORGANIZATION, and LOCATION. 
    \item In MEN-Dataset, we have a total of 101 relations labels. There are 84 relation labels adapted from DocRED and 17 from ACE-05. Meanwhile, DocRED has 96 relation labels. 
    \item MEN-Dataset was developed from news articles while DocRED was developed using Wikipedia documents.
    \item MEN-Dataset consists of news articles with a minimum of four and a maximum of 40 sentences, while the DocRED dataset has a minimum of 2 to a maximum of 20 sentences. The length of the article in DocRED is shorter than MEN-Dataset. 
     \item Most importantly, MEN-Dataset is based on Malaysian English, and DocRED is based on Standard English.
\end{enumerate}
Both datasets feature document-based annotations and encompass both inter- and intra-sentential relations. As there are some differences between the two datasets, we made some modifications in the experiments:
\begin{enumerate}
    \item For entity extraction, we compare the performance of ChatGPT based on entity label PERSON, ORGANIZATION, and LOCATION only. 
    \item For relation extraction, we compare the performance of ChatGPT based on overlapping 84 relations between MEN-Dataset and DocRED.
    \item In the previous section, we evaluated the performance of ChatGPT based on 18 different prompt settings (refer to Appendix \ref{appendix:prompting-techniques}). However, for the DocRED dataset, where the annotation guidelines for entity annotation and explanations for few-shot learning are not available, we specifically applied the following prompting techniques: ZeroShot-NoICL, 3-Iter-ZeroShot-NoICL, 5-Iter-ZeroShot-NoICL, 3Shot-NoICL, and 5Shot-NoICL (refer to Appendix \ref{appendix:prompting-techniques}).
\end{enumerate}

\begin{table}[H]
\centering
\resizebox{\columnwidth}{!}{%
\begin{tabular}{|l|rr|rr|}
\hline
\multicolumn{1}{|c|}{\multirow{2}{*}{\textbf{Prompt Name}}} & \multicolumn{2}{c|}{\textbf{F1-Score (NER)}}                   & \multicolumn{2}{c|}{\textbf{\begin{tabular}[c]{@{}c@{}}F1-Score \\ (Relation Extraction)\end{tabular}}} \\ \cline{2-5} 
\multicolumn{1}{|c|}{}                                      & \multicolumn{1}{l|}{MEN-Dataset} & \multicolumn{1}{l|}{DocRED} & \multicolumn{1}{l|}{MEN-Dataset}                      & \multicolumn{1}{l|}{DocRED}                     \\ \hline
ZeroShot-NoICL                                              & \multicolumn{1}{r|}{0.57}        & 0.65                        & \multicolumn{1}{r|}{0.659}                            & 0.76                                            \\ \hline
3-Iter-ZeroShot-NoICL                                       & \multicolumn{1}{r|}{0.567}       & 0.725                       & \multicolumn{1}{r|}{0.59}                             & 0.654                                           \\ \hline
5-Iter-ZeroShot-NoICL                                       & \multicolumn{1}{r|}{0.558}       & 0.733                       & \multicolumn{1}{r|}{0.58}                             & 0.64                                            \\ \hline
3Shot-NoICL                                                 & \multicolumn{1}{r|}{0.57}        & 0.615                       & \multicolumn{1}{r|}{0.668}                            & 0.663                                           \\ \hline
5Shot-NoICL                                                 & \multicolumn{1}{r|}{0.568}       & 0.738                       & \multicolumn{1}{r|}{0.696}                            & 0.665                                           \\ \hline
\end{tabular}%
}
\caption{Comparing the performance of ChatGPT between MEN-Dataset (Malaysian English) and DocRED (Standard English)}
\label{tab:men-standard-english-chatgpt}
\end{table}

Table \ref{tab:men-standard-english-chatgpt} presents the F1-Scores obtained for this experiment. It is noticeable that the performance of ChatGPT for NER varies significantly between the MEN-Dataset and DocRED datasets. For every prompt setting, the F1-Score for NER in DocRED (Standard English) is higher than MEN-Dataset (Malaysian English). This language-specific performance could be due to the morphosyntactic adaptation that has been discussed and detailed in Section \ref{ssec:eval-chatgpt-on-men-dataset-entity-label}. Meanwhile, the performance of ChatGPT for Relation Extraction does not provide any significant difference between the two datasets. This could be due to the dataset's characteristics, where both were developed for inter- and intra-sentential relations. This result could also be due to morphosyntactic adaptation that can be seen in MEN-Dataset entities only, which does not impact Relation Extraction.  

\section{Conclusion}
\label{sec:conclusion}
In this paper, we comprehensively evaluated and analyzed ChatGPT's ability to extract entities and classify relations from Malaysian English news articles. Our extensive experiment was conducted with 18 different prompting approaches. The experimental results prove that morphosyntactic adaptation impacted the performance of ChatGPT in extracting entities from Malaysian English news articles. We discussed our findings from the experiments, including an analysis of the limitations of ChatGPT.  ChatGPT could not achieve satisfying performance when extracting entities from Malaysian English news articles. Apart from the limitation in understanding the context of inputs, there are a few factors that influenced the performance of ChatGPT. These include the dataset's characteristics, additional contexts like guidelines and explanations, and several few-shot examples. The morphosyntactic adaptation exhibited by Malaysian English influenced the performance of ChatGPT for NER.  Given the annotation of our MEN-Dataset, we could only assess the performance of ChatGPT in NER and RE. For future work, we plan to expand our evaluations by incorporating a broader range of NLP downstream tasks. Furthermore, we will extend our assessment to include other language models, such as GPT-4 \citep{openai_gpt4} and Llama 2 \citep{touvron2023llama}, for NER and RE tasks, specifically in the context of Malaysian English. Finally as a future work, we will also expand the coverage of our experiment with different prompting techniques to ensure our evaluation is statistically significant. 

\section{Ethical Consideration}
\label{sec:ethical-consideration}
In this paper, we evaluated the performance of ChatGPT in extracting entities and relations from Malaysian English news articles. The evaluation was done using news articles (from MEN-Dataset) and Wikipedia articles (from DocRED dataset). No ethics approval was required because these articles were written and published for public consumption. This decision is made after consulting our institution's Human Research Ethics Committee. Besides, ChatGPT was only used to extract information (like entities and relation) from our input and it does not require generating any responses that poses harmful or inappropriate content. As mentioned in Section \ref{ssec:experimental-setup}, we used ChatGPT \footnote{\url{https://chat.openai.com/}} official website and we sent the input one by one, without spamming the website. 

\section{Limitations}
\label{sec:limitation}
Here are some of the limitations in this experiment: 
\begin{enumerate}
    \item As explained in the Introduction (Section \ref{sec:introduction}), various Information Extraction tasks can be done using ChatGPT. However, in this research paper, we focused only on NER and RE due to the annotation of our Malaysian English dataset. In future, we will expand our dataset to cater for other NLP tasks. 
    \item Secondly, we could only conduct the experiments reported in ths paper with small data size. The MEN-Dataset consists of only 200 news articles, with annotated entities and relations. The work on expanding the dataset with more annotated news articles is ongoing, and will be used for thorough experiments and analysis. 
    \item We used ChatGPT Web version instead of OpenAI API in the experiments, due to the following reasons:
    \begin{enumerate}
    \item OpenAI API does not have ability to store information about past interactions. This means, it would have been difficult to provide additional context like Annotation Guideline. However this is not the case when using ChatGPT web interface. LangChain\footnote{\url{https://www.langchain.com/}} has not supported "Memory" functionality when the experiments were conducted. 
    \item Resource Constraint and Efficiency. The utilization of the OpenAI API will incur costs. Small set of data enables better and in-depth analysis ChatGPT outcome.
    \end{enumerate}
\end{enumerate}

\bibliography{anthology,custom}
\bibliographystyle{acl_natbib}

\onecolumn

\appendix




\newpage

\section{Prompt Generated with Entity Annotation Guideline}
\label{appendix:icl-annotation}
\begin{figure*}[h]
    \centering
    \includegraphics[height=\textheight,width=\textwidth,keepaspectratio]{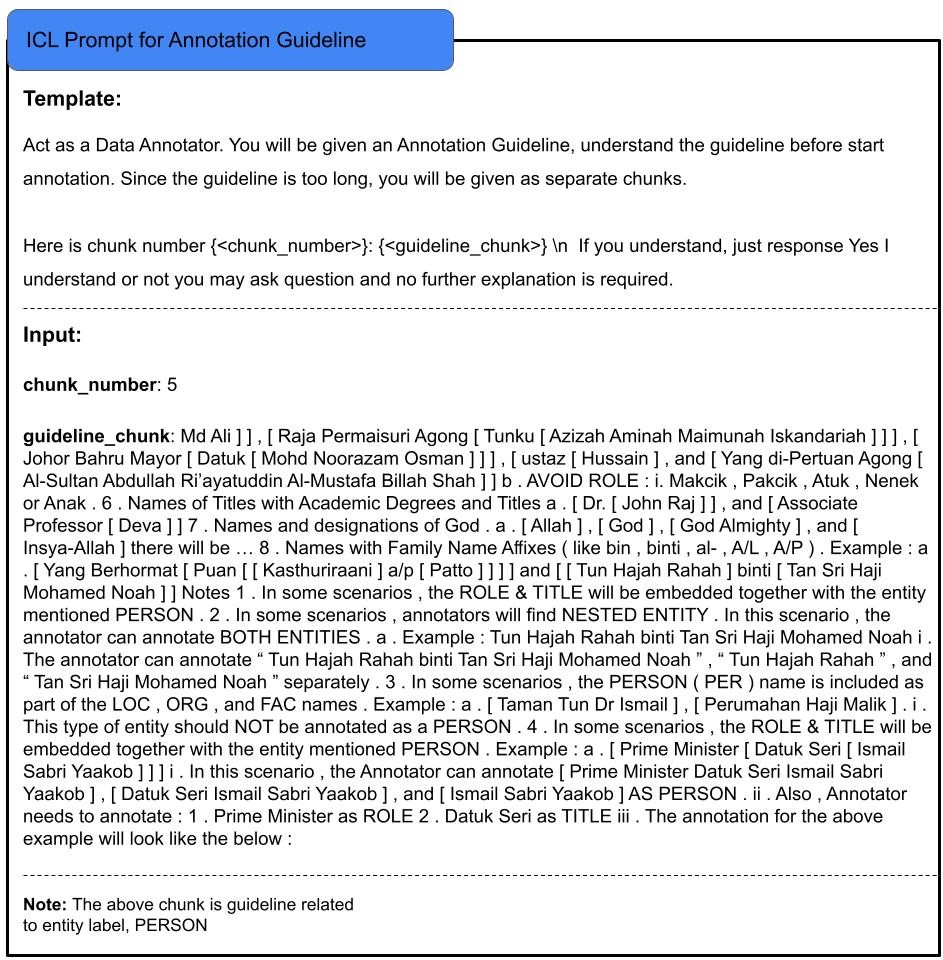}
    \caption{Prompt template used to provide entity annotation guideline as separate chunks}
    \label{fig:icl-annotation}
\end{figure*}

\newpage

\section{Entities and Explanations}
\label{appendix:entities-explainations}
\begin{figure}[H]
    \centering
    \includegraphics[height=\textheight,width=\textwidth,keepaspectratio]{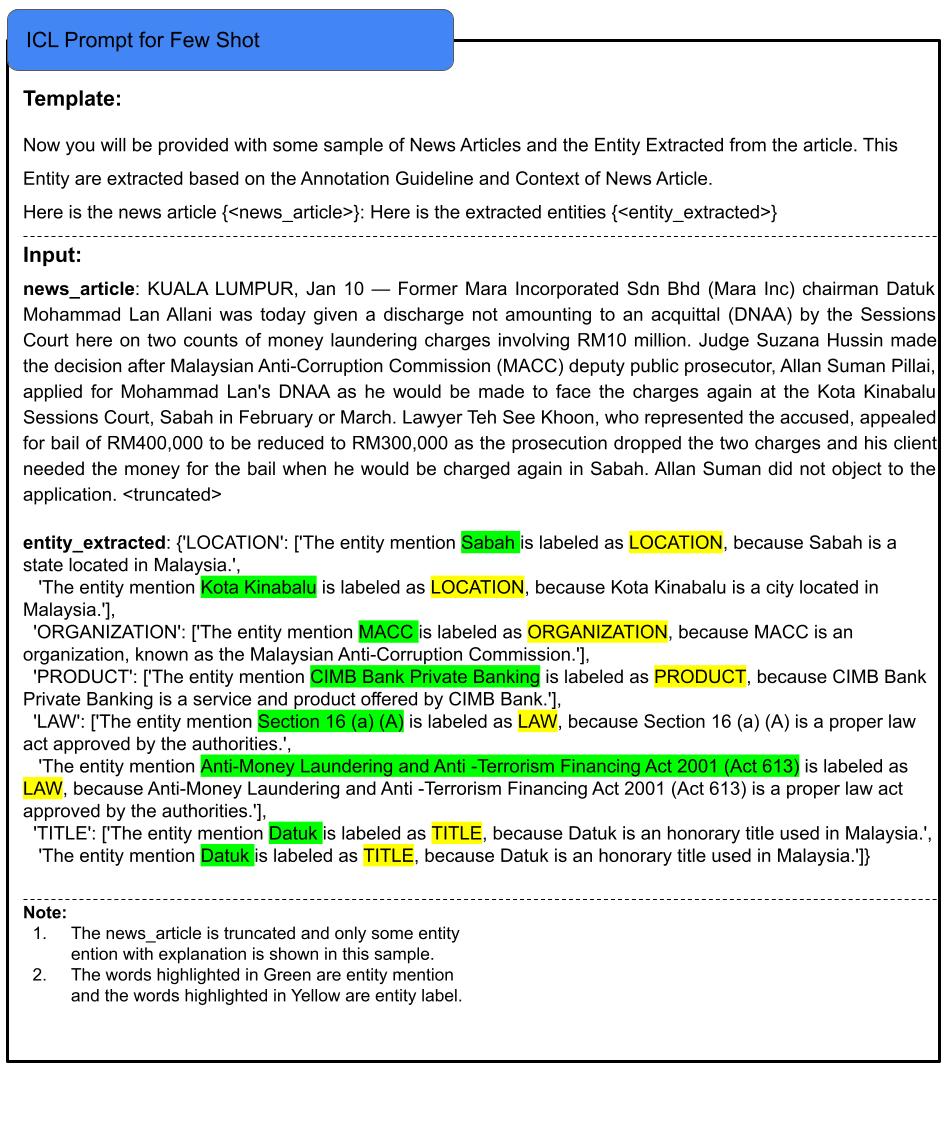}
    \caption{A few examples of manually annotated entities along with explanations for why they have been annotated.}
    \label{fig:entity-explaination}
\end{figure}

\newpage

\section{Prompt for NER Task in ChatGPT}
\label{appendix:ner-task}
\begin{figure}[H]
    \centering
    \includegraphics[height=\textheight,width=\textwidth,keepaspectratio]{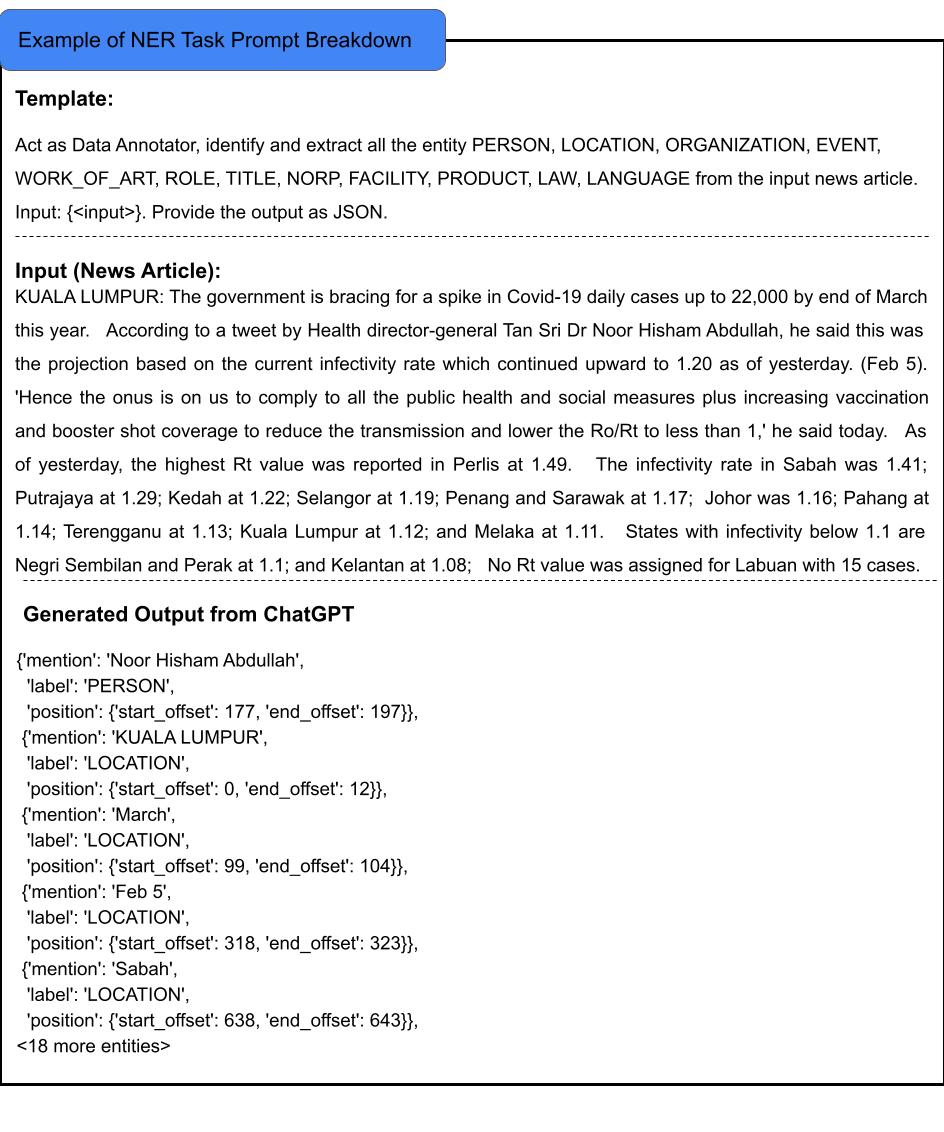}
    \caption{The prompt template used to extract entities based on news article provided.}
    \label{fig:ner-task}
\end{figure}

\newpage 

\section{Prompt for RE Task in ChatGPT}
\label{appendix:re-task}
\begin{figure}[H]
    \centering
    \includegraphics[height=\textheight,width=\textwidth,keepaspectratio]{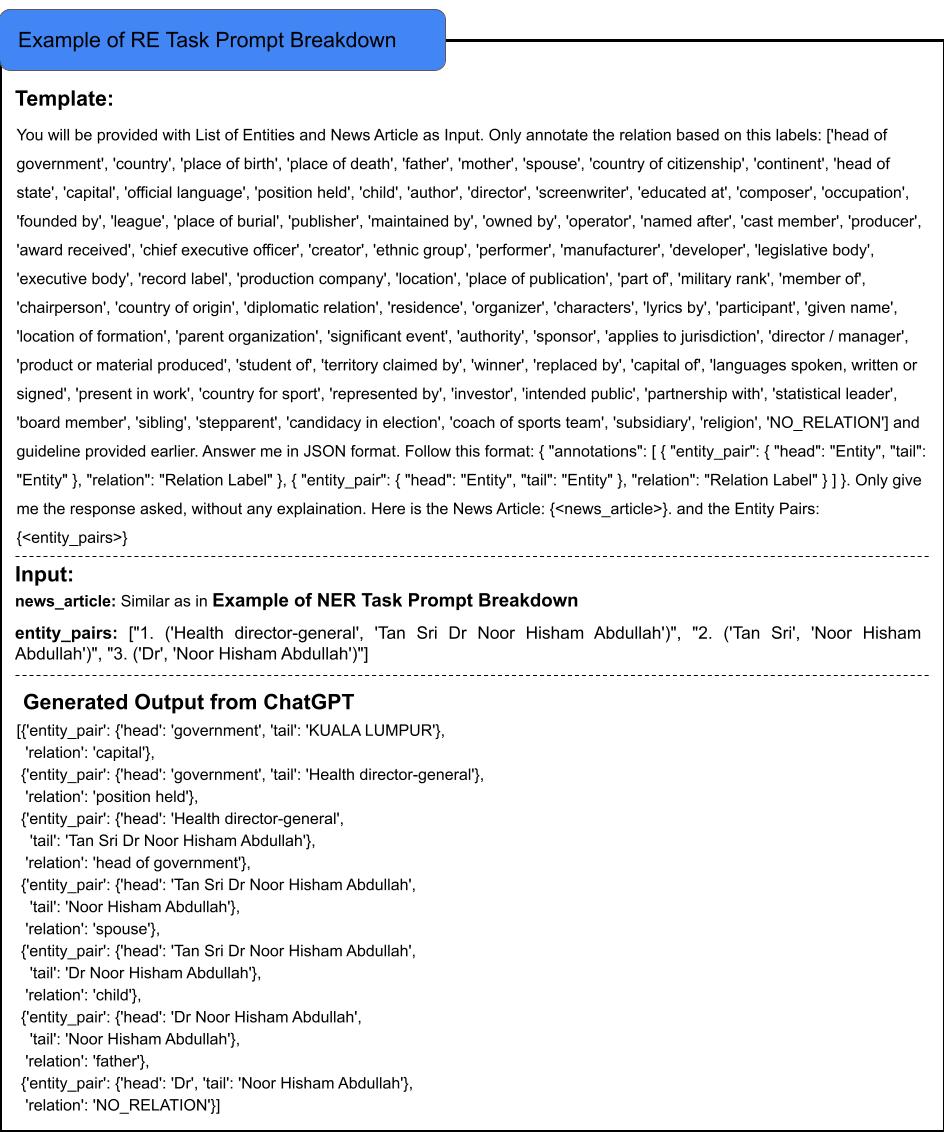}
    \caption{The prompt template used to extract relations based on news article and entities provided.}
    \label{fig:re-task}
\end{figure}

\newpage

\section{List of Named Entity labels}
\label{appendix:men-dataset-entity-labels}
\begin{table}[hbt!]
\centering
\begin{tabular}{|l|l|l|}
\hline
No & Entity Label & Description                                                                                                                                                                                                                                    \\ \hline
1  & PERSON       & \begin{tabular}[c]{@{}l@{}}The Entity PERSON includes Name of Person in the text. \\ This entity type has been adapted from OntoNotes 5.0.\end{tabular}                                                                                        \\ \hline
2  & LOCATION     & \begin{tabular}[c]{@{}l@{}}LOCATION is any place that can be occupied by or has been \\ occupied by someone in this EARTH and outside of EARTH. \\ Entity mention that could be labelled as GPE has been labelled \\ as LOCATION.\end{tabular} \\ \hline
3  & ORGANIZATION & ORGANIZATION is group of people with specific purpose.                                                                                                                                                                                         \\ \hline
4  & NORP         & \begin{tabular}[c]{@{}l@{}}NORP is the abbrevation for the term Nationality, Religious \\ or Political group.\end{tabular}                                                                                                                     \\ \hline
5  & FACILITY     & FACILITY refers to man-made structures.                                                                                                                                                                                                        \\ \hline
6  & PRODUCT      & \begin{tabular}[c]{@{}l@{}}PRODUCT refers to an object, or a service that is made \\ available for consumer use as of the consumer demand.\end{tabular}                                                                                        \\ \hline
7  & EVENT        & \begin{tabular}[c]{@{}l@{}}An EVENT is a reference to an organized or unorganized \\ incident.\end{tabular}                                                                                                                                    \\ \hline
8  & WORK OF ART  & \begin{tabular}[c]{@{}l@{}}WORK OF ART refers to ART entities that has been made \\ by a PERSON or ORGANIZATION.\end{tabular}                                                                                                                  \\ \hline
9  & LAW          & \begin{tabular}[c]{@{}l@{}}LAW are rules that has been made by an authority and that\\ must be obeyed.\end{tabular}                                                                                                                            \\ \hline
10 & LANGUAGE     & LANGUAGE refers to any named language.                                                                                                                                                                                                         \\ \hline
11 & ROLE         & \begin{tabular}[c]{@{}l@{}}ROLE is used to define the position or function of the \\ PERSON in an ORGANIZATION.\end{tabular}                                                                                                                   \\ \hline
12 & TITLE        & TITLE is used to define the honorific title of the PERSON.                                                                                                                                                                                     \\ \hline
\end{tabular}
\caption{Entity Labels}
\label{tab:entity-types}
\centering
\end{table}

\newpage
\begin{landscape}
\section{List of Relation labels}
\label{appendix:men-dataset-relation-labels}
\begin{longtable}{| p{.025\textwidth} | p{.20\textwidth} | p{.10\textwidth} | p{.20\textwidth} | p{.20\textwidth} | p{.60\textwidth} |} 
\hline
No & Relation Label & Dataset Adapted & Entity Type One & Entity Type Two & Description \\ \hline
1 & head of government & DocRED & PER & ORG,LOC & head of the executive power of this town, city, municipality, state, country, or other governmental body \\ \hline
2 & country & DocRED & PER,ORG & LOC & sovereign state of this item (not to be used for human beings) \\ \hline
3 & place of birth & DocRED & PER & LOC & most specific known (e.g. city instead of country, or hospital instead of city) birth location of a person, animal or fictional character \\ \hline
4 & place of death & DocRED & PER & LOC & most specific known (e.g. city instead of country, or hospital instead of city) death location of a person, animal or fictional character \\ \hline
5 & father & DocRED & PER & PER & "male parent of the subject." \\ \hline
6 & mother & DocRED & PER & PER & "female parent of the subject." \\ \hline
7 & spouse & DocRED & PER & PER & "the subject has the object as their spouse (husband, wife, partner, etc.)." \\ \hline
8 & country of citizenship & DocRED & LOC & PER & the object is a country that recognizes the subject as its citizen \\ \hline
9 & continent & DocRED & LOC & LOC & continent of which the subject is a part \\ \hline
10 & head of state & DocRED & PER & LOC & official with the highest formal authority in a country/state \\ \hline
11 & capital & DocRED & LOC & LOC & seat of government of a country, province, state or other type of administrative territorial entity \\ \hline
12 & official language & DocRED & LOC,ORG & PER & language designated as official by this item \\ \hline
13 & position held & DocRED & PER & ROLE & subject currently or formerly holds the object position or public office \\ \hline
14 & child & DocRED & PER & PER & subject has object as child. Do not use for stepchildren \\ \hline
15 & author & DocRED & PER & WORK\_OF\_ART & main creator(s) of a written work \\ \hline
16 & director & DocRED & PER & WORK\_OF\_ART & director(s) of film, TV-series, stageplay, video game or similar \\ \hline
17 & screenwriter & DocRED & PER & WORK\_OF\_ART & person(s) who wrote the script for subject item \\ \hline
18 & educated at & DocRED & PER & ORG & educational institution attended by subject \\ \hline
19 & composer & DocRED & PER & WORK\_OF\_ART & "person(s) who wrote the music" \\ \hline
20 & occupation & DocRED & PER & ROLE & "occupation of a person" \\ \hline
21 & founded by & DocRED & PER & ORG & founder or co-founder of this organization, religion or place \\ \hline
22 & league & DocRED & ORG & EVENT & league in which team or player plays or has played in \\ \hline
23 & place of burial & DocRED & PER & LOC & location of grave, resting place, place of ash-scattering, etc. (e.g., town/city or cemetery) for a person or animal. There may be several places: e.g., re-burials, parts of body buried separately. \\ \hline
24 & publisher & DocRED & PER & WORK\_OF\_ART & organization or person responsible for publishing books, periodicals, printed music, podcasts, games or software \\ \hline
25 & maintained by & DocRED & PER,ORG & FAC,ORG & person or organization in charge of keeping the subject (for instance an infrastructure) in functioning order \\ \hline
26 & owned by & DocRED & PER & ORG, FAC, PRODUCT & owner of the subject \\ \hline
27 & operator & DocRED & PER & PRODUCT,FAC & person, profession, or organization that operates the equipment, facility, or service \\ \hline
28 & named after & DocRED & PER & FAC,ORG,EVENT & "entity or event that inspired the subject's name, or namesake (in at least one language)." \\ \hline
29 & cast member & DocRED & PER & WORK\_OF\_ART & "actor in the subject production" \\ \hline
30 & producer & DocRED & PER & WORK\_OF\_ART & person(s) who produced the film, musical work, theatrical production, etc. (for film, this does not include executive producers, associate producers, etc.) \\ \hline
31 & award received & DocRED & PER, ORG, WORK\_OF\_ART, TITLE & WORK\_OF\_ART, TITLE & award or recognition received by a person, organization or creative work \\ \hline
32 & chief executive officer & DocRED & PER & ORG & highest-ranking corporate officer appointed as the CEO within an organization \\ \hline
33 & creator & DocRED & PER & WORK\_OF\_ART, PRODUCT & maker of this creative work or other object (where no more specific property exists) \\ \hline
34 & ethnic group & DocRED & PER & ORG & subject's ethnicity (consensus is that a VERY high standard of proof is needed for this field to be used. In general this means 1) the subject claims it themselves, or 2) it is widely agreed on by scholars, or 3) is fictional and portrayed as such) \\ \hline
35 & performer & DocRED & PER & WORK\_OF\_ART & actor, musician, band or other performer associated with this role or musical work \\ \hline
36 & manufacturer & DocRED & ORG & PRODUCT & manufacturer or producer of this product \\ \hline
37 & developer & DocRED & ORG,PER & PRODUCT,FAC & organization or person that developed the item \\ \hline
38 & legislative body & DocRED & ORG & ORG & legislative body governing this entity; political institution with elected representatives, such as a parliament/legislature or council \\ \hline
39 & executive body & DocRED & ORG & ORG & branch of government for the daily administration of the territorial entity \\ \hline
40 & record label & DocRED & ORG & WORK\_OF\_ART & brand and trademark associated with the marketing of subject music recordings and music videos \\ \hline
41 & production company & DocRED & ORG & WORK\_OF\_ART & company that produced this film, audio or performing arts work \\ \hline 
42 & location & DocRED & PER,FAC,ORG & LOC & location of the object, structure or event. \\ \hline
43 & place of publication & DocRED & WORK\_OF\_ART & LOC & geographical place of publication of the edition (use 1st edition when referring to works) \\ \hline
44 & part of & DocRED & PER & ORG,EVENT & "object of which the subject is a part (if this subject is already part of object A which is a part of object B, then please only make the subject part of object A)." \\ \hline
45 & military rank & DocRED & PER & ROLE & "military rank achieved by a person (should usually have a ""start time"" qualifier), or military rank associated with a position" \\ \hline
46 & member of & DocRED & PER & ORG & organization, club or musical group to which the subject belongs. Do not use for membership in ethnic or social groups, nor for holding a political position, such as a member of parliament. \\ \hline
47 & chairperson & DocRED & PER & ORG & presiding member of an organization, group or body \\ \hline
48 & country of origin & DocRED & LOC & WORK\_OF\_ART, PRODUCT & country of origin of this item (creative work, food, phrase, product, etc.) \\ \hline
49 & diplomatic relation & DocRED & ORG & ORG & diplomatic relations of the country \\ \hline
50 & residence & DocRED & PER & FAC,LOC & the place where the person is or has been, resident \\ \hline
51 & organizer & DocRED & PER,ORG & EVENT & person or institution organizing an event \\ \hline
52 & characters & DocRED & PER & WORK\_OF\_ART & characters which appear in this item (like plays, operas, operettas, books, comics, films, TV series, video games) \\ \hline
53 & lyrics by & DocRED & PER & WORK\_OF\_ART & author of song lyrics \\ \hline
54 & participant & DocRED & PER,ORG & EVENT,ORG & "person, group of people or organization (object) that actively takes/took part in an event or process (subject)." \\ \hline
55 & given name & DocRED & PER & PER & first name or another given name of this person; values used with the property should not link disambiguations nor family names \\ \hline
56 & location of formation & DocRED & ORG & LOC & location where a group or organization was formed \\ \hline
57 & parent organization & DocRED & ORG & ORG & parent organization of an organization. \\ \hline
58 & significant event & DocRED & PER,ORG & EVENT & significant or notable events associated with the subject \\ \hline
59 & authority & DocRED & PER & ORG & entity having executive power on given entity \\ \hline
60 & sponsor & DocRED & PER,ORG & PER,EVENT & organization or individual that sponsors this item \\ \hline
61 & applies to jurisdiction & DocRED & LAW & LOC & the item (institution, law, public office, public register...) or statement belongs to or has power over or applies to the value (a territorial jurisdiction: a country, state, municipality, ...) \\ \hline
62 & director / manager & DocRED & PER & ORG & person who manages any kind of group \\ \hline
63 & product or material produced & DocRED & PER & WORK\_OF\_ART & material or product produced by a government agency, business, industry, facility, or process \\ \hline
64 & student of & DocRED & PER & PER & person who has taught this person \\ \hline
65 & territory claimed by & DocRED & ORG & LOC & administrative divisions that claim control of a given area \\ \hline
66 & winner & DocRED & PER,ORG & EVENT & "winner of a competition or similar event, not to be used for awards or for wars or battles" \\ \hline
67 & replaced by & DocRED & PER & PER & "other person or item which continues the item by replacing it in its role." \\ \hline
68 & capital of & DocRED & LOC & LOC & country, state, department, canton or other administrative division of which the municipality is the governmental seat \\ \hline
69 & languages spoken, written or signed & DocRED & PER & LANGUAGE & language(s) that a person or a people speaks, writes or signs, including the native language(s) \\ \hline
70 & present in work & DocRED & PER & WORK\_OF\_ART & this (fictional or fictionalized) entity or person appears in that work as part of the narration \\ \hline
71 & country for sport & DocRED & PER,ORG & LOC & country a person or a team represents when playing a sport \\ \hline
72 & represented by & DocRED & PER & ORG & person or agency that represents or manages the subject \\ \hline
73 & investor & DocRED & PER,ORG & ORG & individual or organization which invests money in the item for the purpose of obtaining financial return on their investment \\ \hline
74 & intended public & DocRED & PER,ORG & PRODUCT,EVENT & this work, product, object or event is intended for, or has been designed to that person or group of people, animals, plants, etc \\ \hline
75 & partnership with & DocRED & ORG & ORG & partnership (commercial or/and non-commercial) between this organization and another organization or institution \\ \hline
76 & statistical leader & DocRED & ORG,PER & EVENT & leader of a sports tournament in one of statistical qualities (points, assists, rebounds etc.). \\ \hline
77 & board member & DocRED & PER & ORG & member(s) of the board for the organization \\ \hline
78 & sibling & DocRED & PER & PER & "the subject and the object have at least one common parent (brother, sister, etc. including half-siblings)" \\ \hline
79 & stepparent & DocRED & PER & PER & subject has the object as their stepparent \\ \hline
80 & candidacy in election & DocRED & PER,ORG & EVENT & election where the subject is a candidate \\ \hline
81 & coach of sports team & DocRED & PER & ORG & sports club or team for which this person is or was on-field manager or coach \\ \hline
82 & subsidiary & DocRED & ORG & ORG & subsidiary of a company or organization; generally a fully owned separate corporation. \\ \hline
83 & religion & DocRED & PER & ORG & religion of a person, organization or religious building, or associated with this subject \\ \hline
84 & Physical.Located & ACE-2005 & PER & FAC, LOC & Located captures the physical location of an entity. \\ \hline
85 & Physical.Near & ACE-2005 & PER, FAC, LOC & FAC, LOC & Indicates that an entity is explicitly near another entity. \\ \hline
86 & Part-Whole.Geo & ACE-2005 & FAC, LOC & FAC, LOC & Captures the location of FAC, LOC, or GPE in or at or as a part of another FAC, LOC or GPE. \\ \hline
87 & Part-Whole.Subsidary & ACE-2005 & ORG & ORG, LOC & Captures the ownership, administrative, and other hierarchical relationships between organizations and between organizations and GPEs. \\ \hline
88 & Per-Social.Business & ACE-2005 & PER & PER & Captures the connection between two entities in any professional relationships. \\ \hline
89 & Per-Social.Family & ACE-2005 & PER & PER & Captures the connection between one entity and another entity in family relations. \\ \hline
90 & Per-Social.Lasting & ACE-2005 & PER & PER & Captures the relations that invlovle personal contact (Where one entity has spent time with another entity, like classmate, neighbor), or indication that the relationships exists outside of a particular cited interaction. \\ \hline
91 & Org-Aff.Employment & ACE-2005 & PER & ORG,LOC & Captures relationship between Person and their employers. \\ \hline
92 & Org-Aff.Ownership & ACE-2005 & PER & ORG & Captures relationship between a Person and an Organization owned by that PERSON \\ \hline
93 & Org-Aff.Founder & ACE-2005 & PER,ORG & ORG,LOC & Captures relation between an entity and an organization that has been founded by the entity \\ \hline
94 & Org-Aff.Student-Alum & ACE-2005 & PER & ORG-Educational ONLY & Captures relation between Person and an educational institution. \\ \hline
95 & Org-Aff.Sports-Affiliation & ACE-2005 & PER & ORG & Captures relation between Player, Coach, Manager with their affiliated Sport ORG \\ \hline
96 & Org-Aff.Shareholder & ACE-2005 & PER, ORG, GPE & ORG, GPE & Captures the relation between an agent and an Organization \\ \hline
97 & Org-Aff.Membership & ACE-2005 & PER, ORG, GPE & ORG & Membership captures relation between an entity and organization which the entity is a member of \\ \hline
98 & Agent-Artifact.UOIM & ACE-2005 & PER, ORG, GPE & FAC & When an entity own an artifact, uses an artifact or caused an artifact to come into being. \\ \hline
99 & Gen-Aff.CRRE & ACE-2005 & PER & ORG, LOC & "When there is a relation between PER and LOC in which they have citizenship. Or when there is a relation between PER and LOC they live. Or when when there is a relation between PER and religious ORG or PER. Or when there is a relation between PER and LOC or PER entity that indicates their ethnicity"  \\ \hline
100 & Gen-Aff.Loc-Origin & ACE-2005 & ORG & LOC & Captures the relation between an organization and the LOC where it is located. \\ \hline
101 & NO\_RELATION & ~ & ANY ENTITY & ANY ENTITY & Can be used for any entity pair that does not have a suitable Relations Listed \\ \hline
\end{longtable}
\end{landscape}



\newpage

\begin{landscape}
\section{Different Prompting Techniques}
\label{appendix:prompting-techniques}
\begin{longtable}{|p{3cm}|p{4cm}|p{4cm}|l|}
\hline
Prompt Name & Prompt Technique & ICL   & Description                                                                                                                                                                                                                                                                                                                                                                                                                                                                                   \\ \hline
\endfirsthead
\endhead
ZeroShot-NoICL& Zero Shot                                                  & None                  & \begin{tabular}[c]{@{}l@{}}Only news articles will be given to ChatGPT. Based on the \\ existing knowledge, ChatGPT will need to extract entities \\ and relation.\end{tabular}                                                                                                                                                                                                                                                                                                               \\ \hline
ZeroShot-Guide& Zero Shot                                                  & Guideline             & \begin{tabular}[c]{@{}l@{}}Only annotation guideline will be provided to ChatGPT. \\ ChatGPT will need to extract entities and relation based \\ on guideline.\end{tabular}                                                                                                                                                                                                                                                                                                                   \\ \hline
3-Iter-ZeroShot-NoICL& Self Consistent Zero Shot (3 Iteration)                    & None                  & \begin{tabular}[c]{@{}l@{}}Only provide news articles to ChatGPT. No additional \\ context will be given. Based on the existing knowledge, \\ ChatGPT will need to extract entities and relation.\end{tabular}                                                                                                                                                                                                                                                                                \\ \hline
5-Iter-ZeroShot-NoICL& Self Consistent Zero Shot (5 Iteration)                    & None                  & \begin{tabular}[c]{@{}l@{}}No additional context will be given to ChatGPT. The entity\\ or relation that is consistently extract from similar news \\ article will selected as final output.\end{tabular}                                                                                                                                                                                                                                                                                     \\ \hline
3-Iter-ZeroShot-Guide& Self Consistent Zero Shot (3 Iteration)                    & Guideline             & \begin{tabular}[c]{@{}l@{}}Annotation guideline will be given to ChatGPT. The entity\\ or relation that is consistently extract from similar news \\ article will selected as final output.\end{tabular}                                                                                                                                                                                                                                                                                      \\ \hline
5-Iter-ZeroShot-Guide& Self Consistent Zero Shot (5 Iteration)                    & Guideline             & \begin{tabular}[c]{@{}l@{}}Annotation guideline will be given to ChatGPT. The entity\\ or relation that is consistently extract from similar news \\ article will selected as final output.\end{tabular}                                                                                                                                                                                                                                                                                      \\ \hline
3Shot-NoICL& 3 - Shot Learning                                          & None                  & \begin{tabular}[c]{@{}l@{}}Three news articles with entities and relation extracted \\ will given as context to ChatGPT. ChatGPT will need \\ to extract entities and relation based existing knowledge and\\ provided sample news articles.\end{tabular}                                                                                                                                                                                                                                     \\ \hline
3Shot-Guide& 3 - Shot Learning                                          & Guideline             & \begin{tabular}[c]{@{}l@{}}Together with three news articles, ChatGPT will be \\ provided with annotation guideline. ChatGPT will need to \\ extract entities and relation based existing knowledge and\\ provided sample news articles.\end{tabular}                                                                                                                                                                                                                                         \\ \hline
3Shot-Explain& 3 - Shot Learning                                          & Explaination          & \begin{tabular}[c]{@{}l@{}}Each instance of an entity and relation will be accompanied \\ by an explanation for its extraction. ChatGPT's task will \\ involve extracting entities and relations using the existing \\ knowledge and information provided in the sample news\\ articles.\end{tabular}                                                                                                                                                                                         \\ \hline
3Shot-Guide\_Explain& 3 - Shot Learning                                          & Guideline+Explanation & \begin{tabular}[c]{@{}l@{}}Each instance of an entity and relation will be accompanied \\ by an explanation for its extraction. Additionally, the annotation\\ guideline will also be give to ChatGPT. ChatGPT's task will \\ involve extracting entities and relations using the existing \\ knowledge and information provided in the sample news\\ articles.\end{tabular}                                                                                                                  \\ \hline
3-Iter-3Shot-Guide\_Explain& Self Consistent Sampling (3 Iteration) + 3 - Shot Learning & Guideline+Explanation & \begin{tabular}[c]{@{}l@{}}Each instance of an entity and relation will be accompanied \\ by an explanation for its extraction. Additionally, the \\ annotation guideline will also be give to ChatGPT. ChatGPT's \\ task will involve extracting entities and relations using the \\ existing knowledge and information provided in the sample\\ news articles. The entity or relation that is consistently extract \\ from similar news article will selected as final output.\end{tabular} \\ \hline
5-Iter-3Shot-Guide\_Explain& Self Consistent Sampling (5 Iteration) + 3 - Shot Learning & Guideline+Explanation & \begin{tabular}[c]{@{}l@{}}Each instance of an entity and relation will be accompanied \\ by an explanation for its extraction. Additionally, the \\ annotation guideline will also be give to ChatGPT. ChatGPT's \\ task will involve extracting entities and relations using the \\ existing knowledge and information provided in the sample\\ news articles. The entity or relation that is consistently extract \\ from similar news article will selected as final output.\end{tabular} \\ \hline
5Shot-NoICL& 5 - Shot Learning                                          & None                  & The explaination is similar to 3 - Shot Learning.                                                                                                                                                                                                                                                                                                                                                                                                                                             \\ \hline
5Shot-Guide& 5 - Shot Learning                                          & Guideline             & The explaination is similar to 3 - Shot Learning.                                                                                                                                                                                                                                                                                                                                                                                                                                             \\ \hline
5Shot-Explain& 5 - Shot Learning                                          & Explaination          & The explaination is similar to 3 - Shot Learning.                                                                                                                                                                                                                                                                                                                                                                                                                                             \\ \hline
5Shot-Guide\_Explain& 5 - Shot Learning                                          & Guideline+Explanation & The explaination is similar to 3 - Shot Learning.                                                                                                                                                                                                                                                                                                                                                                                                                                             \\ \hline
3-Iter-5Shot-Guide\_Explain& Self Consistent Sampling (3 Iteration) + 5 - Shot Learning & Guideline+Explanation & The explaination is similar to 3 - Shot Learning.                                                                                                                                                                                                                                                                                                                                                                                                                                             \\ \hline
5-Iter-5Shot-Guide\_Explain& Self Consistent Sampling (5 Iteration) + 5 - Shot Learning & Guideline+Explanation & The explaination is similar to 3 - Shot Learning.                                                                                                                                                                                                                                                                                                                                                                                                                                             \\ \hline
\caption{Different prompting techniques used to evaluate ChatGPT capabilities for NER and Relation Extraction}
\label{tab:prompt-ner-re}\\
\end{longtable}
\end{landscape}

\newpage


\newpage

\begin{landscape}
\section{Evaluating ChatGPT NER Capability with MEN-Dataset (From Perspective of Entity Label)}
\label{appendix:eval-ner-men-entity-label}
\begin{table}[H]
\centering
\resizebox{\columnwidth}{!}{%
 & 0.511                                                                                          & 0.607                                                                                            & 0.609                                                                                                & 0.221                                                                                       & 0.247                                                                                           & 0.09                                                                                          & 0.366                                                                                        & 0                                                                                                 & 0                                                                                             & 0.474                                                                                     & 0.36                                                                                        & 0.038                                                                                        \\ \hline
   & Average F1-Score                                                       & 0.507                                                                                          & 0.616                                                                                            & 0.61                                                                                                 & 0.218                                                                                       & 0.212                                                                                           & 0.132                                                                                         & 0.382                                                                                        & 0.005                                                                                             & 0                                                                                             & 0.427                                                                                     & 0.305                                                                                       & 0.035                                                                                        \\ \hline
\end{tabular}%
}
\caption{The F1-Score from the perspective of entity label.}
\label{tab:f1-score-entity-label}
\end{table}
\end{landscape}

\end{document}